\documentclass[journal,english]{IEEEtran}
\usepackage{url}
\usepackage{acronym}
\usepackage{amsmath, amsthm, amssymb, amsfonts, mathtools}
\usepackage{babel}
\usepackage{xcolor}
\usepackage{float}
\usepackage{graphicx}
\graphicspath{{img/}}
\usepackage[T1]{fontenc}
\usepackage{tablefootnote}
\usepackage{units}
\usepackage{siunitx}
\usepackage{verbatim}
\usepackage{times}
\usepackage{mathtools}
\usepackage{float}
\usepackage{caption}
\usepackage{subcaption}
\usepackage{cite}
\usepackage{multirow}
\usepackage{gensymb}
\usepackage{tabularx}
\usepackage{booktabs}
\usepackage{csquotes}

\usepackage{balance}

\usepackage{xargs}
\usepackage{mdframed}

\usepackage[noabbrev,capitalize]{cleveref}

\usepackage[colorinlistoftodos,prependcaption,textsize=scriptsize]{todonotes}
\newcommandx{\note}[2][1=]
{\todo[inline,linecolor=black,backgroundcolor=green!25,bordercolor=green,#1]{#2}}

\title{Embodied Neuromorphic Vision\\ with Event-Driven Random Backpropagation}

\author{
  \noindent Jacques Kaiser$^1$, Alexander Friedrich$^1$, J. Camilo Vasquez Tieck$^1$, \\
  \noindent Daniel Reichard$^1$, Arne Roennau$^1$, Emre Neftci$^2$, R\"udiger Dillmann$^1$
\thanks{$^1$FZI Research Center For Information Technology, Karlsruhe, Germany}
\thanks{$^2$Department of Cognitive Sciences, University of California Irvine, Irvine, USA}}

\begin{document}


\markboth{}%
{Shell \MakeLowercase{\textit{Kaiser et al.}}: Embodied random backprop}

\maketitle

\acrodef{AC}[AC]{Arrenhius \& Current}
\acrodef{ANN}[ANN]{Artificial Neural Network}
\acrodef{AER}[AER]{Address Event Representation}
\acrodef{AEX}[AEX]{AER EXtension board}
\acrodef{AMDA}[AMDA]{``AER Motherboard with D/A converters''}
\acrodef{API}[API]{Application Programming Interface}
\acrodef{BPTT}[BPTT]{Back-Propagation-Through-Time}
\acrodef{BP}[BP]{Back-Propagation}
\acrodef{BM}[BM]{Boltzmann Machine}
\acrodef{CAVIAR}[CAVIAR]{Convolution AER Vision Architecture for Real-Time}
\acrodef{CCN}[CCN]{Cooperative and Competitive Network}
\acrodef{CD}[CD]{Contrastive Divergence}
\acrodef{CMOS}[CMOS]{Complementary Metal--Oxide--Semiconductor}
\acrodef{COTS}[COTS]{Commercial Off-The-Shelf}
\acrodef{CPU}[CPU]{Central Processing Unit}
\acrodef{CV}[CV]{Coefficient of Variation}
\acrodef{CV}[CV]{Coefficient of Variation}
\acrodef{DAC}[DAC]{Digital--to--Analog}
\acrodef{DBN}[DBN]{Deep Belief Network}
\acrodef{DCLL}[DCLL]{Deep Continuous Local Learning}
\acrodef{EP}[EP]{Eligibility Propagation}
\acrodef{DFA}[DFA]{Deterministic Finite Automaton}
\acrodef{DFA}[DFA]{Deterministic Finite Automaton}
\acrodef{divmod3}[DIVMOD3]{divisibility of a number by 3}
\acrodef{DPE}[DPE]{Dynamic Parameter Estimation}
\acrodef{DPI}[DPI]{Differential-Pair Integrator}
\acrodef{DSP}[DSP]{Digital Signal Processor}
\acrodef{DVS}[DVS]{Dynamic Vision Sensor}
\acrodef{EDVAC}[EDVAC]{Electronic Discrete Variable Automatic Computer}
\acrodef{EIF}[EI\&F]{Exponential Integrate \& Fire}
\acrodef{EIN}[EIN]{Excitatory--Inhibitory Network}
\acrodef{EPSC}[EPSC]{Excitatory Post-Synaptic Current}
\acrodef{EPSP}[EPSP]{Excitatory Post--Synaptic Potential}
\acrodef{eRBP}[eRBP]{Event-Driven Random Back-Propagation}
\acrodef{FPGA}[FPGA]{Field Programmable Gate Array}
\acrodef{FSM}[FSM]{Finite State Machine}
\acrodef{GPU}[GPU]{Graphical Processing Unit}
\acrodef{HAL}[HAL]{Hardware Abstraction Layer}
\acrodef{HH}[H\&H]{Hodgkin \& Huxley}
\acrodef{HMM}[HMM]{Hidden Markov Model}
\acrodef{HW}[HW]{Hardware}
\acrodef{hWTA}[hWTA]{Hard Winner--Take--All}
\acrodef{IF2DWTA}[IF2DWTA]{Integrate \& Fire 2--Dimensional WTA}
\acrodef{IF}[I\&F]{Integrate \& Fire}
\acrodef{IFSLWTA}[IFSLWTA]{Integrate \& Fire Stop Learning WTA}
\acrodef{INCF}[INCF]{International Neuroinformatics Coordinating Facility}
\acrodef{INI}[INI]{Institute of Neuroinformatics}
\acrodef{IO}[IO]{Input-Output}
\acrodef{IPSC}[IPSC]{Inhibitory Post-Synaptic Current}
\acrodef{ISI}[ISI]{Inter--Spike Interval}
\acrodef{JFLAP}[JFLAP]{Java - Formal Languages and Automata Package}
\acrodef{LIF}[LI\&F]{Linear Integrate \& Fire}
\acrodef{LSM}[LSM]{Liquid State Machine}
\acrodef{LTD}[LTD]{Long-Term Depression}
\acrodef{LTI}[LTI]{Linear Time-Invariant}
\acrodef{LTP}[LTP]{Long-Term Potentiation}
\acrodef{LTU}[LTU]{Linear Threshold Unit}
\acrodef{MCMC}{Markov Chain Monte Carlo}
\acrodef{NHML}[NHML]{Neuromorphic Hardware Mark-up Language}
\acrodef{NMDA}[NMDA]{NMDA}
\acrodef{NME}[NE]{Neuromorphic Engineering}
\acrodef{PCB}[PCB]{Printed Circuit Board}
\acrodef{PRC}[PRC]{Phase Response Curve}
\acrodef{PSC}[PSC]{Post-Synaptic Current}
\acrodef{PSP}[PSP]{Post--Synaptic Potential}
\acrodef{RI}[KL]{Kullback-Leibler}
\acrodef{RRAM}[RRAM]{Resistive Random-Access Memory}
\acrodef{RBM}[RBM]{Restricted Boltzmann Machine}
\acrodef{ROC}[ROC]{Receiver Operator Characteristic}
\acrodef{SAC}[SAC]{Selective Attention Chip}
\acrodef{SCD}[SCD]{Spike-Based Contrastive Divergence}
\acrodef{SCX}[SCX]{Silicon CorteX}
\acrodef{SRM}[SRM]{Spike Response Model}
\acrodef{SNN}[SNN]{Spiking Neural Network}
\acrodef{STDP}[STDP]{Spike Time Dependent Plasticity}
\acrodef{SW}[SW]{Software}
\acrodef{sWTA}[SWTA]{Soft Winner--Take--All}
\acrodef{VHDL}[VHDL]{VHSIC Hardware Description Language}
\acrodef{VLSI}[VLSI]{Very  Large  Scale  Integration}
\acrodef{WTA}[WTA]{Winner--Take--All}
\acrodef{XML}[XML]{eXtensible Mark-up Language}

\begin{abstract}
  Spike-based communication between biological neurons is sparse and unreliable.
  This enables the brain to process visual information from the eyes efficiently.
  Taking inspiration from biology, artificial spiking neural networks coupled with silicon retinas attempt to model these computations.
  Recent findings in machine learning allowed the derivation of a family of powerful synaptic plasticity rules approximating backpropagation for spiking networks.
  Are these rules capable of processing real-world visual sensory data?
  In this paper, we evaluate the performance of \ac{eRBP} at learning representations from event streams provided by a \ac{DVS}.
  First, we show that \ac{eRBP} matches state-of-the-art performance on the DvsGesture dataset with the addition of a simple covert attention mechanism.
  By remapping visual receptive fields relatively to the center of the motion, this attention mechanism provides translation invariance at low computational cost compared to convolutions.
  Second, we successfully integrate \ac{eRBP} in a real robotic setup, where a robotic arm grasps objects according to detected visual affordances.
  In this setup, visual information is actively sensed by a \ac{DVS} mounted on a robotic head performing microsaccadic eye movements.
  We show that our method classifies affordances within 100ms after microsaccade onset, which is comparable to human performance reported in behavioral study.
  Our results suggest that advances in neuromorphic technology and plasticity rules enable the development of autonomous robots operating at high speed and low energy consumption.
\end{abstract}

\begin{IEEEkeywords}
Neurorobotics, Spiking Neural Networks, Event-Based Vision, Learning, Synaptic Plasticity
\end{IEEEkeywords}

\IEEEpeerreviewmaketitle

\section{Introduction}
\acresetall

\IEEEPARstart{T}{here} are important discrepancies between the computational paradigms of the brain and modern computer architectures.
Remarkably, evolution discovered a way to learn and perform energy-efficient computations from large networks of neurons. 
Particularly, the communication of spikes between neurons is asynchronous, sparse and unreliable.
Understanding what computations the brain performs and how they are implemented on a neural substrate is a global endeavor of our time.
From an engineering perspective, this would enable the design of autonomous learning robots operating at high speed for a fraction of the energy budget of current solutions.

\begin{figure}[!htbp]
  \centering
  \includegraphics[width=\columnwidth]{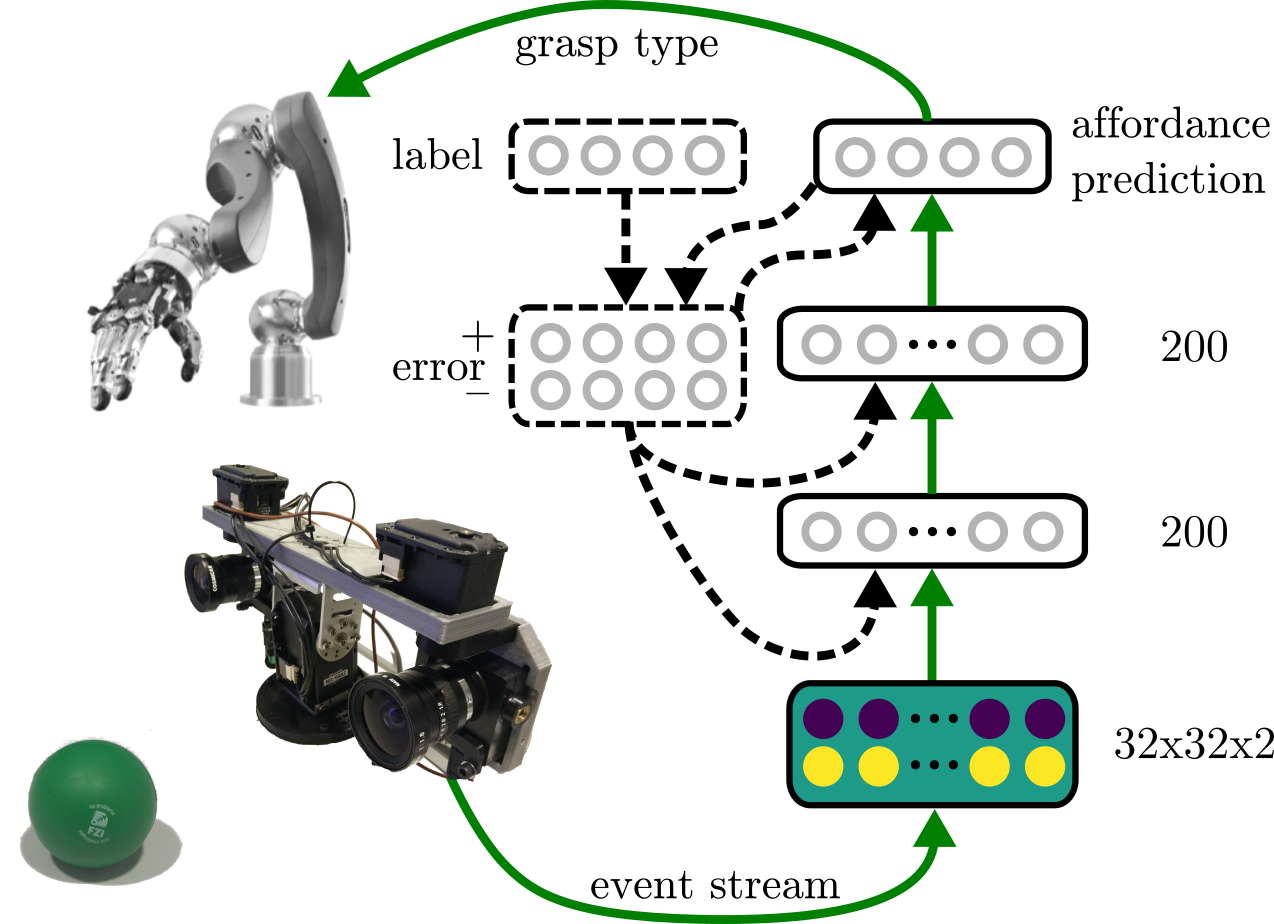}
  \caption{
    Our robotic setup embodying the synaptic learning rule \ac{eRBP} \cite{neftci2017event}.
    The \ac{DVS} is mounted on a robotic head performing microsaccadic eye movements.
    The spiking network is trained (dashed-line connections) in a supervised fashion to classify visual affordances from the event streams online.
    During training, output neurons and label neurons project to an error neuron population, which conveys a random feedback to hidden layers.
    The output neurons of the network correspond to the four types of affordances: ball-grasp, bottle-grasp, pen-grasp or do nothing.
    At test time, a Schunk LWA4P arm equipped with a five-finger Schunk SVH gripper performs the detected reaching and grasping motion.
  }
  \label{fig:network}
\end{figure}

Recently, a family of synaptic learning rules building on groundbreaking machine learning research have been proposed in \cite{neftci2017event,Zenke_Ganguli17_supesupe,kaiser2018synaptic,bellec2019biologically}.
These rules implement variations of \ac{BP} adapted to spiking networks by approximating the gradient computations.
As was shown in \cite{lillicrap2016random,Jaderberg_etal16}, approximations of the gradient can be made without significantly deteriorating the accuracy of the network.
These approximations allow weight updates to only depend on local information available at the synapse, enabling online learning with an efficient neuromorphic hardware implementation.
Are these rules capable of efficiently processing complex visual information like the brain?
Synaptic learning rules are rarely evaluated in an embodiment, and often report below state-of-the-art performance compared to classical methods \cite{bing2018survey}.


In this paper, we evaluate the ability of one of these rules -- \ac{eRBP} \cite{neftci2017event} -- to learn from real visual event-based sensor data.
\ac{eRBP} was the first synaptic plasticity rule formally derived from Random Backpropagation \cite{lillicrap2016random}, its deep learning equivalent.
First, we benchmark \ac{eRBP} on the popular visual event-based dataset DvsGesture \cite{Amir2017} from IBM.
We introduce a covert attention mechanism which further improves the performance by providing translation invariance without convolutions.
Second, we integrate the rule in a real-world closed-loop robotic grasping setup involving a robotic head, arm and a five-finger hand.
The spiking network learns to classify different types of affordances based on visual information obtained with microsaccades, and communicates this information to the arm for grasping.
This real-world task has the potential to enhance neuromorphic and neurorobotics research since many functional components such as reaching, grasping and depth perception can be easily segregated and implemented with brain models.
This work paves the way towards the integration of brain-inspired computational paradigms into the field of robotics.

This paper is structured as follows.
We give a brief introduction to the synaptic learning rule \ac{eRBP} in \cref{sec:approach} together with the network architecture.
We further present our covert attention mechanism and the microsaccadic eye movements.
Our approach is evaluated in \cref{sec:eval}, first on the DvsGesture \cite{Amir2017} benchmark, second in our real-world grasping setup.
We conclude in \cref{sec:conc}.

\section{Approach}\label{sec:approach}

\subsection{Event-Driven Random Backpropagation}\label{sec:erbp}



\ac{eRBP} \cite{neftci2017event} is an interpretation for spiking neurons of feedback alignment presented in \cite{lillicrap2016random} for analog neurons.
Feedback alignment is an approximation of backpropagation, where the prediction error is fed back to all neurons directly with fixed, random feedback weights.
With a mean-square error loss, the update for a hidden weight from neuron $j$ to neuron $i$ can be formulated as follows:
\begin{equation}\label{eq:lr}
  \begin{split}
    \Delta w_{ij}(t) &= y_j \phi' \left(\sum_{j'} w_{ij'}(t) y_{j'}(t)\right) T_i(t), \\
    T_i(t) &= \sum_k e_k(t) g_{ik},
  \end{split}
\end{equation}
\noindent with $y_j$ the output of neuron $j$, $\phi$ the activation function of neuron $i$, $e_k$ the prediction error for output neuron $k$ and $g_{ik}$ a fixed random feedback weight.
This rule is interpreted for spiking neurons with $y_j$ the spiking output ($0$ or $1$) of neuron $j$ and $\phi$ the function mapping membrane potential to spikes.
The weighted sum of input spikes $\sum_{j'} w_{ij'}(t) y_{j'}(t)$ is interpreted as the membrane potential $I_i$ of neuron $i$.
\ac{eRBP} relies on hard-threshold Leaky Integrate-And-Fire neurons, leading $\phi$ to be non-differentiable.
Following the approach of surrogate gradients \cite{neftci2019surrogate}, $\phi'$ is approximated as the boxcar function, equal to $1$ between $b_\text{min}$ and $b_\text{max}$ and $0$ otherwise.
Note that the temporal dynamics of the Leaky Integrate-And-Fire neuron -- post-synaptic potentials and refractory period -- are not taken into account in \cref{eq:lr}.
More recent plasticity rules now include an eligibility trace \cite{Zenke_Ganguli17_supesupe,kaiser2018synaptic,bellec2019biologically} accounting for some of these dynamics.

This rule can be efficiently implemented in a spiking network with weight updates triggered by pre-synaptic spikes $y_j$ as:
\begin{equation}\label{eq:deltaw}
  \Delta w_{ij}(t) \propto \begin{cases}
    -\sum_{k} e_k(t)g_{ik} & \mbox{if } b_\text{min} < I_i < b_\text{max} \\
    0 & \mbox{otherwise}
  \end{cases},
\end{equation}
\noindent with $b_\text{min}$ and $b_\text{max}$ the window of the boxcar function.
The weight update for the last layer is an exception, since there are no random feedback weights.
Each time a pre-synaptic neuron $j$ with a connection to an output neuron $k$ spikes, the weight update of this connection is calculated by:
\begin{equation}
  \Delta w_{kj}(t) \propto \begin{cases}
    -e_k(t) & \mbox{if } b_\text{min} < I_k < b_\text{max} \\
    0 & \mbox{otherwise}
  \end{cases}.
\end{equation}
\noindent Since \ac{eRBP} only uses two comparisons and one addition for each pre-synaptic spike to perform the weight update, it allows a real-time, energy-efficient and online learning implementation running on neuromorphic hardware.

\subsection{Network Architecture} \label{sec:network}

We propose a similar network architecture to the one proposed in \cite{neftci2017event}.
Specifically, the network consists of spiking neurons organized in feedforward layers, performing classification with one-hot encoding.
In other words, the $k$ output neurons of the last layer correspond to the $k$ class of the dataset.
The error signals $e_k(t)$ are encoded in spikes and provided to all hidden neurons with backward connections from error neurons.
These error spikes are integrated in a dedicated leaking dendritic compartment locally for every learning neuron.
The weight of the backward connections are drawn from a random distribution and fixed -- they correspond to the factors $g_{ik}$ in \cref{eq:lr,eq:deltaw}.

For each class, there are two error neurons conveying positive and negative error respectively.
This is required, since spikes are not signed events.
A pair of positive and negative error neurons are connected to the corresponding output neuron and label neuron.
There is one label neuron for each class, which is spiking repeatedly during training when a sample of the respective class is presented.
Formally, the error $e_k(t)$ is approximated as:
\begin{equation}
  \begin{aligned}
    e_k(t) & \cong \nu_k^{+}(t)-\nu_k^{-}(t),\\
    \nu_k^{+}(t) & \propto \nu_k^P(t)-\nu_k^L(t),\\
    \nu_k^{-}(t) & \propto -\nu_k^P(t)+\nu_k^L(t),
    \end{aligned}
\end{equation}
\noindent where $\nu_k^P(t), \nu_k^L(t)$ are the firing rates of the output neurons and label neurons respectively, and $\nu_k^{+}(t), \nu_k^{-}(t)$ are the firing rates of the positive and negative error neurons.
Within this framework, all computations are performed with spiking neurons and communicated as spikes, including the computation of errors.
As in the brain, all synapses are stochastic, with a probability of dropping spikes.
The network is depicted in \cref{fig:network}.

In this work, the network learns from event streams provided by a \ac{DVS}.
Since spikes are not signed events, we associate two neurons for each pixel to convey ON- and OFF-events separately.
This distinction is important since event polarities carry the instantaneous information of direction of motion (see \cref{fig:arm_circling}).
Since events are generated only upon light change, two different setup are analyzed: a dataset where changes originate from motion in the scene, and a dataset where changes originate from fixational eye movements.
The evaluation on these two types of dataset is important since they can lead to different performance \cite{kaiser2018synaptic}.

\subsection{Covert Attention Window} \label{sec:attention}

\begin{figure}[!htbp]
    \centering
    \begin{subfigure}[t]{0.45\columnwidth}
        \centering
        \includegraphics[width=\columnwidth]{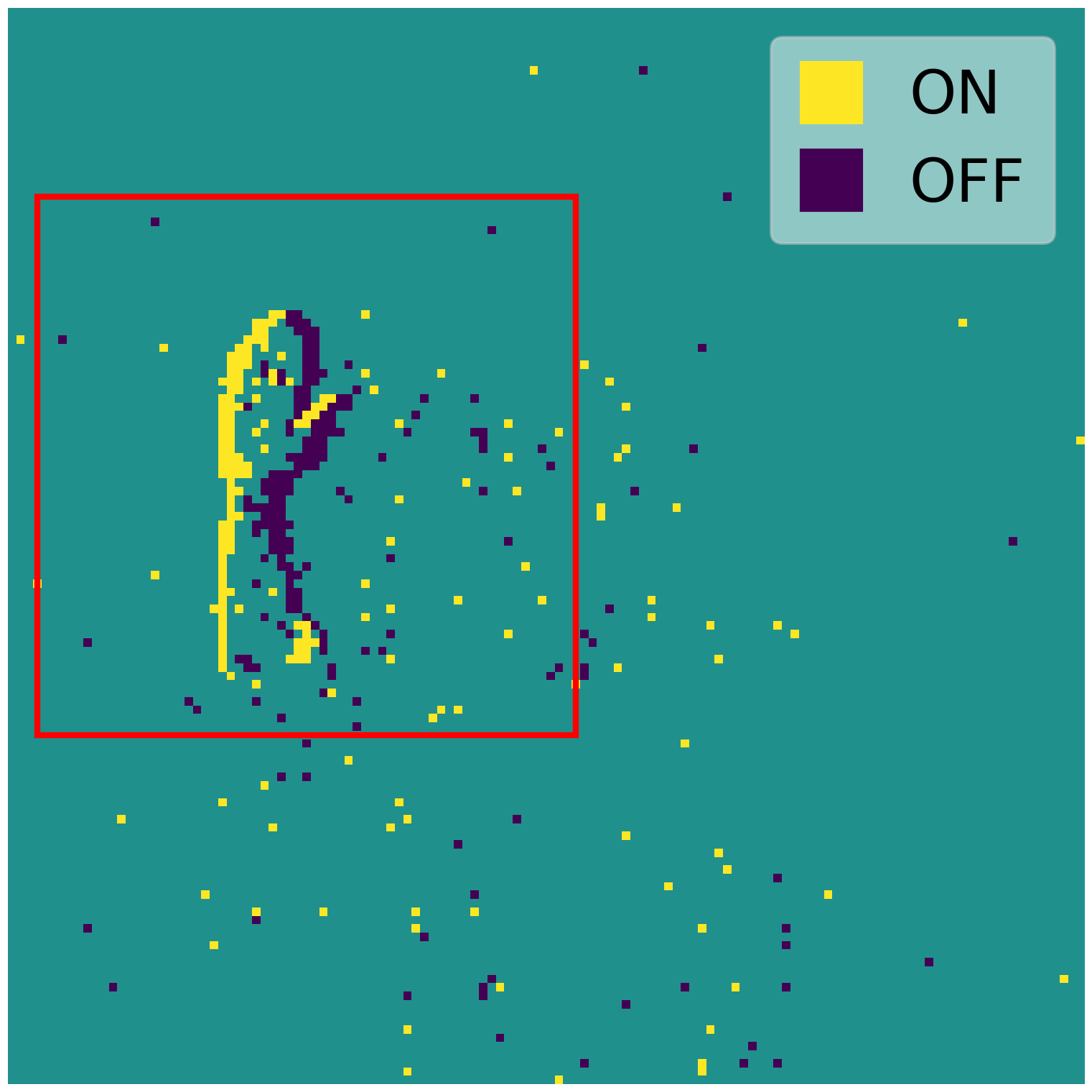}
        \caption{Arm circling clockwise}
        \label{fig:arm_circling_clock}
    \end{subfigure}%
    ~
    \begin{subfigure}[t]{0.45\columnwidth}
        \centering
        \includegraphics[width=\columnwidth]{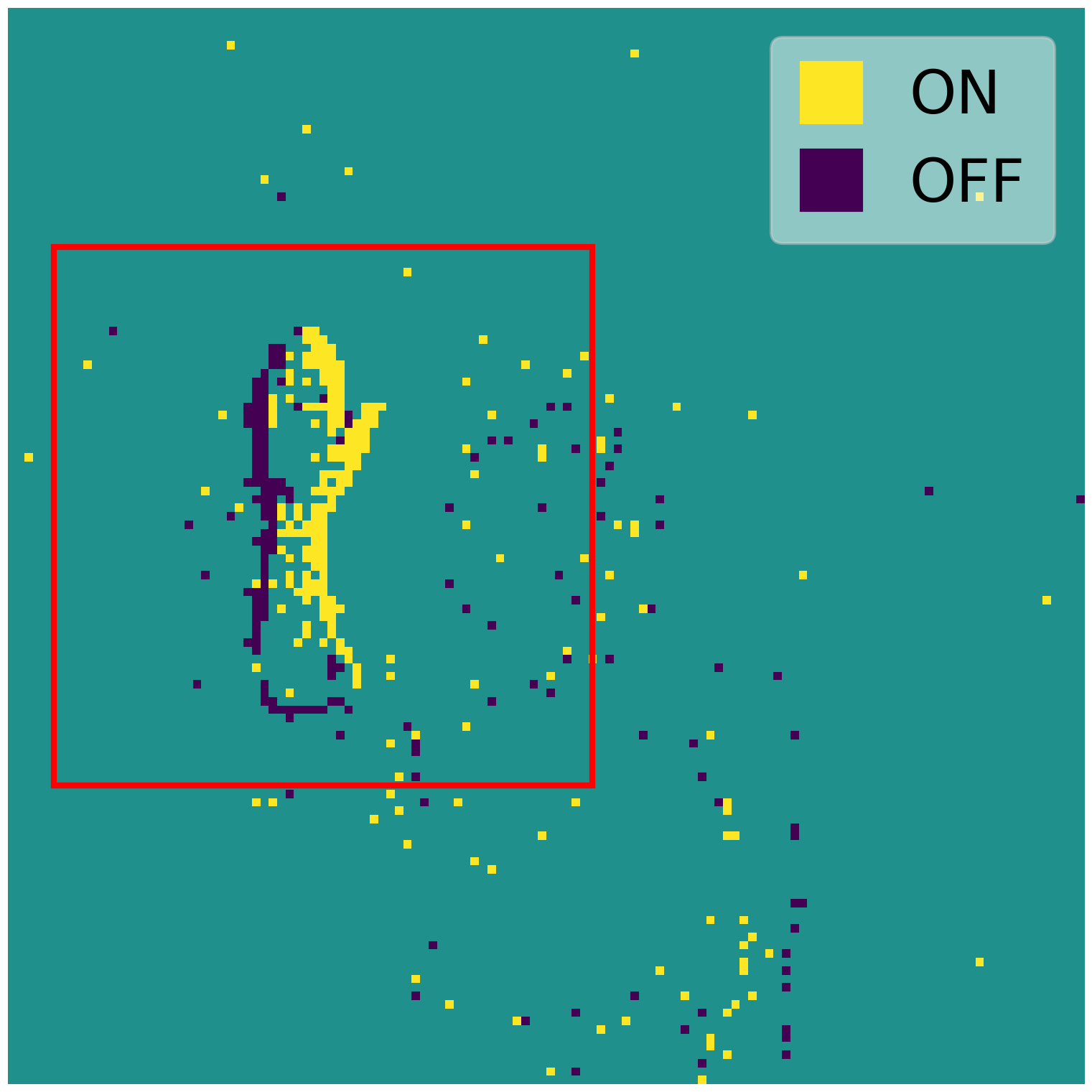}
        \caption{Arm circling anticlockwise}
        \label{fig:arm_circling_anticlock}
    \end{subfigure}
    \caption{
      Aggregation of 1000 events for two samples of the DvsGesture dataset (user 10).
      Yellow pixels symbolize ON-events, blue pixels are OFF-events.
      The information about direction of motion is contained in the event polarity, hence the importance of their segregation in the input layer.
      The red square represents the attention window of size $64 \times 64$, calculated as the median of the last 1000 events.
    }
    \label{fig:arm_circling}
\end{figure}

It was shown in biology that visual receptive fields of frontal eye field neurons are constantly remapped \cite{zirnsak2014visual,sommer2006influence}.
Inspired from this insight, we introduce a simple covert attention mechanism which consists of moving an attention window across the input stream.
Covert attention, as opposed to overt attention, signifies an attention shift which was not marked by eye movements.
Particularly suited to event streams, the center of the attention window is computed online as the median address event of the last $n_\text{attention}$ events, see \cref{fig:arm_circling}.
By remapping receptive fields relatively to the center of the motion, this technique enables translation invariance at low computational cost compared to convolutions.
This method also allows to reduce the dimension of the event stream without rescaling.

A similar method was already introduced in \cite{Zhao2014} for classifying a dataset of three human motions (bend, sit/stand, walk) recorded with a \ac{DVS}.
Their approach consists of remapping the address of their feature neurons (C1) with respect to their mean activation before being fed to the classifier.
Instead, our method consists of remapping the address events directly, with respect to the median event.
Unlike the median, the mean activation can result in an event-less attention window in case of multiple objects in motion, such as two-hand gestures.
Additionally, since our attention window is smaller than the event stream, eccentric events are not processed by the network.
We show in this paper how this biologically motivated technique boosts the performance, even on DvsGesture, where multiple body parts are simultaneously in motion.
We note that a similar mechanism could be integrated in a robotic head as the one used in this paper to perform actual eye movements (see \cref{fig:microsaccade}).
However, an additional mechanism to discard events resulting of the ego-motion would be required, which could be based on visual prediction \cite{kaiser2017scaling,kaiser2018il,sommer2006influence}.

\subsection{Microsaccadic eye-movements}\label{sec:microsaccade}

\begin{figure}[!htbp]
  \centering
  \includegraphics[width=0.7\columnwidth]{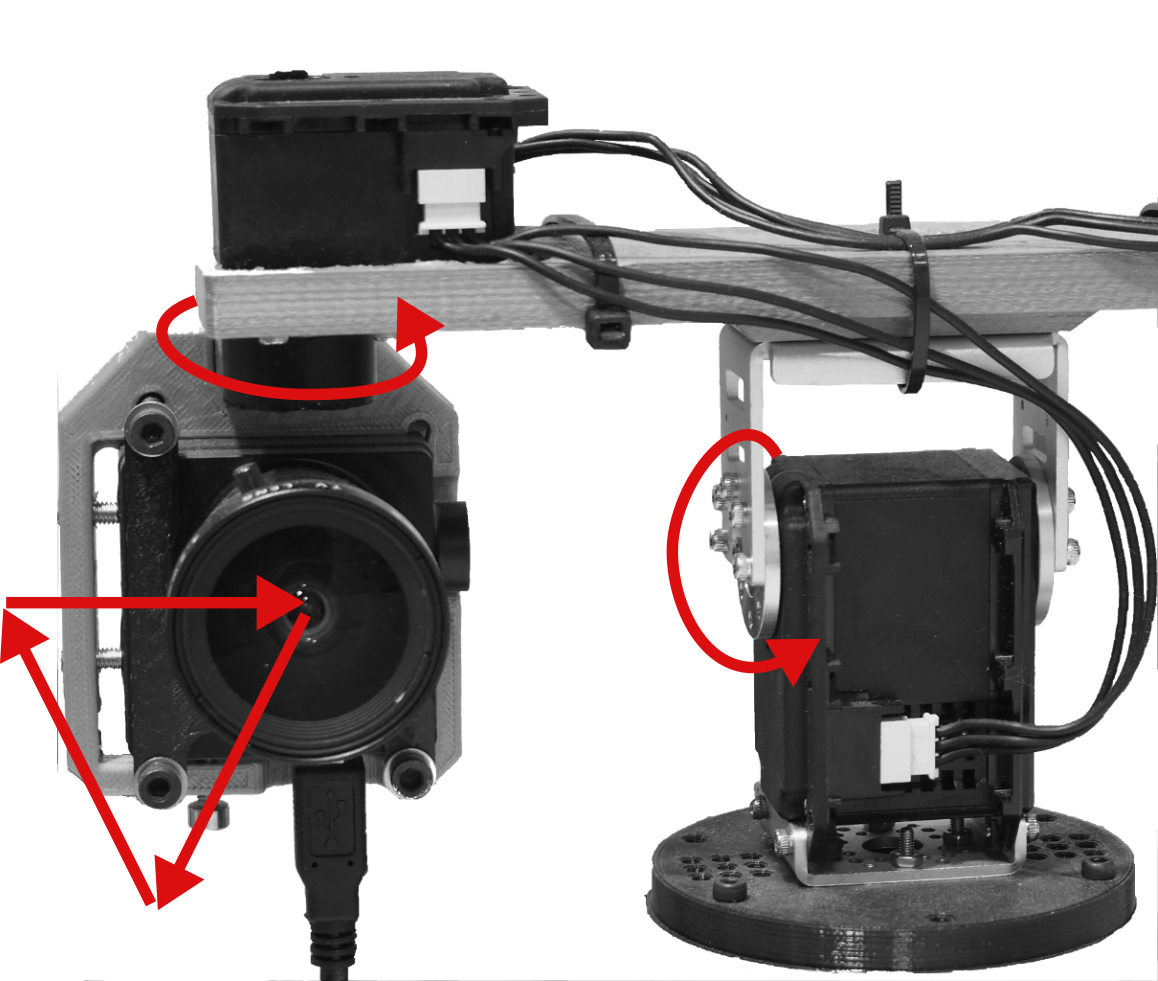}
  \caption{
    Microsaccadic motion of the \ac{DVS} performed by the robotic head.
    The motion consists of three phases.
    An negative tilt of $\alpha$ and negative pan of $\alpha/2$,
    followed by a tilt of $\alpha$ and negative pan of $\alpha/2$,
    followed by a final pan of $\alpha$ moving the \ac{DVS} back to its initial position.
    We chose the angle $\alpha=\SI{1.833}{\degree}$.
    Each motion is effectuated in \SI{0.2}{\second}.
  }
  \label{fig:microsaccade}
\end{figure}

For our real-world grasping experiment, address events are sensed from static scenes by performing microsaccadic eye movements.
This technique was already used to convert images to event streams \cite{Orchard2015a}, essentially extracting edge features \cite{kaiser2018microsaccades}.
To this end, we mounted the \ac{DVS} on the robotic head presented in \cite{kaiser2018microsaccades}, see \cref{fig:microsaccade}.
One Dynamixel servo MX-64AT is used to tilt both \ac{DVS} simultaneously, while two other Dynamixel servos MX-28AT are used to pan each \ac{DVS} independently.
The center of all rotations is approximately the optical center of each \ac{DVS}.
In this work, only the events of the right \ac{DVS} are processed.
The microsaccadic motion consists of an isosceles triangle in joint space.
The three motions defined by the edges of the triangle last \SI{0.2}{\second} each.
The first motion consists of a negative tilt of $\alpha$ and negative pan of $\alpha/2$.
The second motion is a tilt of $\alpha$ and negative pan of $\alpha/2$.
The third motion moves the \ac{DVS} back to its initial position with a pan of $\alpha$.
We chose the angle $\alpha=\SI{1.833}{\degree}$.
This angle is much smaller in biology, but \ac{DVS} pixels are much larger than the photoreceptors of the retina \cite{Martinez-Conde2004}.

The microsaccades are triggered either manually for recording training data, or automatically in a loop at test time.
We allow the events to flow through the network only when a microsaccade is triggered.
No information about the properties of the microsaccade is passed as an input to the network.

\section{Evaluation}\label{sec:eval}

Throughout the experiments, we rely on a dense 4-layers architecture with two hidden layers of 200 neurons respectively (see \cref{fig:network}).
As described in \cref{sec:network}, the ON- and OFF-events obtained from the \ac{DVS} are segregated in two separate input layers.
All synapses are stochastic, with a 35\% chance of dropping each spike.
All the presented experiments relied on the open-source implementation of \ac{eRBP}\footnote{\url{https://gitlab.com/eneftci/erbp_auryn}} based on the neural simulator Auryn \cite{zenke2014limits}.

\subsection{DvsGesture}

\begin{figure}[!htbp]
  \centering
  \includegraphics[width=\columnwidth]{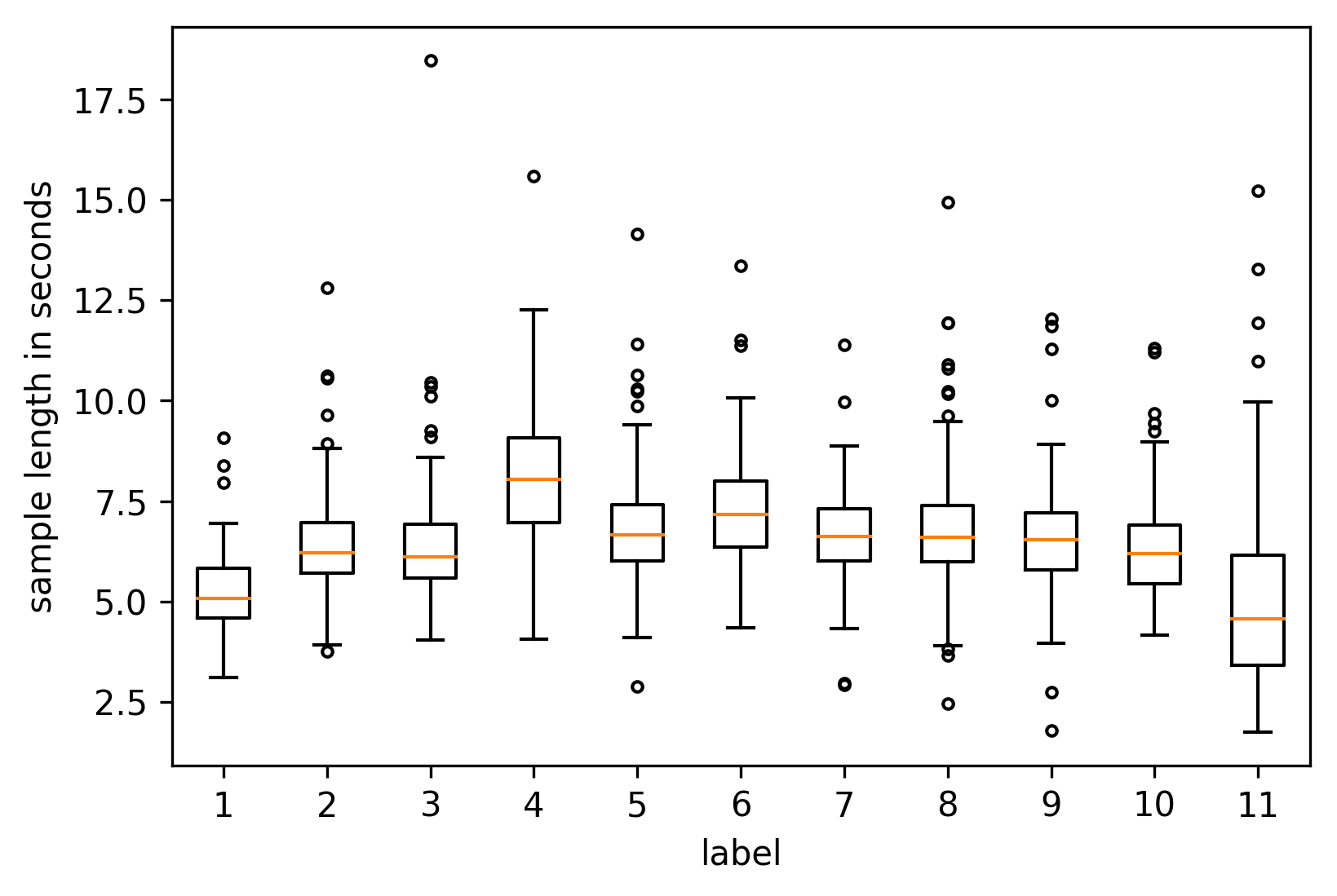}
  \caption{
    Statistics on sample duration for each label in the DvsGesture dataset.
    The red horizontal lines represent the median sample duration.
    Each box indicates the interquartile range ($IQR = Q_3 - Q_1$) per label, which is the range between the first $Q_1$ and the third quartile $Q_3$.
    Whisker pairs show the range of all sample durations within $Q1-1.5\times IQR$ and $Q3+1.5\times IQR$.
    Outliers are represented by small circles.
  }
  \label{fig:sample_lengths}
\end{figure}

DvsGesture is an action recognition dataset recorded by IBM using a \ac{DVS} \cite{Amir2017,Lichtsteiner2008}.
We reduce the dimensionality of the input event stream to $64 \times 64$, both in the rescaling case and the covert attention window case.
It consists of 29 subjects performing 11 diverse actions in three different illumination conditions.
We split the whole dataset into 1176 training samples and 288 test samples, leaving out the data when users do not perform a labeled motion.
This training set consists of 7602s (approximately 2h) of recordings, versus 1960s for the test set.
The duration of the actions varies greatly across samples, see \cref{tab:action_rec_labels} and \cref{fig:sample_lengths}.
A single sample may be about 1 second or over 18 seconds long.
Despite this high variance, all samples were used in full without temporal modifications, both for training and testing.
The number of events to calculate the position of the attention window was set to $n_\text{attention}=1000$.

\begin{table}[h]
  \centering
  \begin{tabularx}{\columnwidth}{@{}cccX@{}}
    \toprule
    Label & \#Training & \#Test & Description\\
    \midrule
    1 & 97 & 24 & Clapping\\
    2 & 98 & 24 & Right hand waving\\
    3 & 98 & 24 & Left hand waving\\
    4 & 98 & 24 & Right arm circling clockwise\\
    5 & 98 & 24 & Right arm circling anticlockwise\\
    6 & 98 & 24 & Left arm circling clockwise\\
    7 & 99 & 24 & Left arm circling anticlockwise\\
    8 & 196 & 48 & Arms rolling\\
    9 & 98 & 24 & Air drum\\
    10 & 98 & 24 & Air guitar\\
    11 & 98 & 24 & Other gestures\\
    \bottomrule
  \end{tabularx}
  \caption{
    Labels of all the different classes in the DvsGesture dataset and their description.
    The amount of samples of label 8 is doubled, since arms rolling were recorded and labeled with 8 for both rotation directions.}
  \label{tab:action_rec_labels}
\end{table}

\begin{figure}[!htbp]
\centering
\includegraphics[width=\columnwidth]{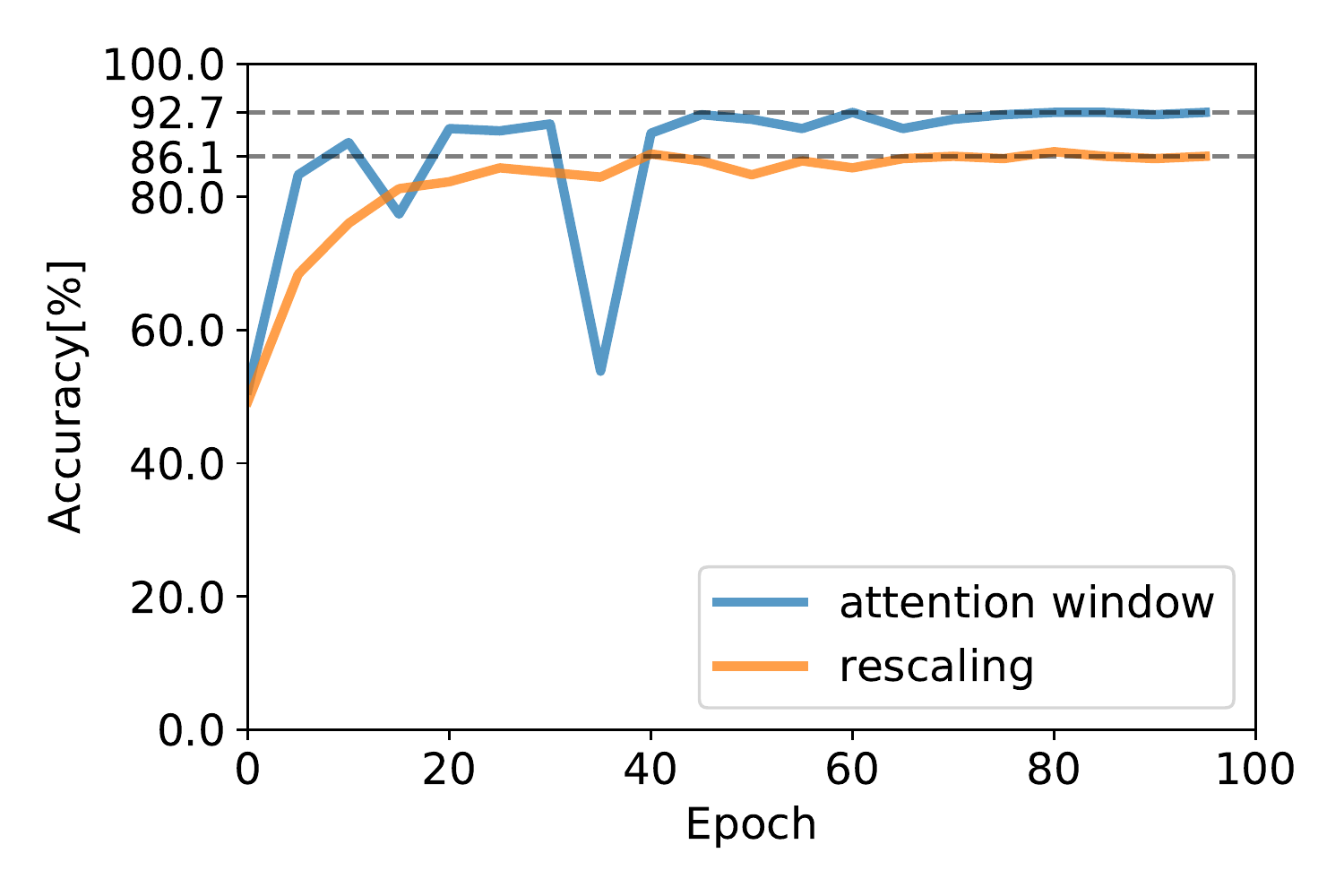}
\caption{
  Classification accuracy over 100 epochs of training on DvsGesture.
  The accuracy converge to 86.1\% and 92.7\% for the rescaling and the attention window approaches respectively.
  The dimension of the event stream is $64 \times 64$ in both cases.
  The drop in accuracy in early stages of learning in the attention window case are due to the stochastic synapses, which have 35\% chances of dropping spikes.
  The rescaling approach is resilient to this stochasticity since all events in the original stream are squeezed into macro-pixels, leading to redundant events.
}
\label{fig:exp78_77_79:acc_hist}
\end{figure}

\begin{figure*}[!hbtp]
    \begin{center}
      \includegraphics[width=.9\textwidth]{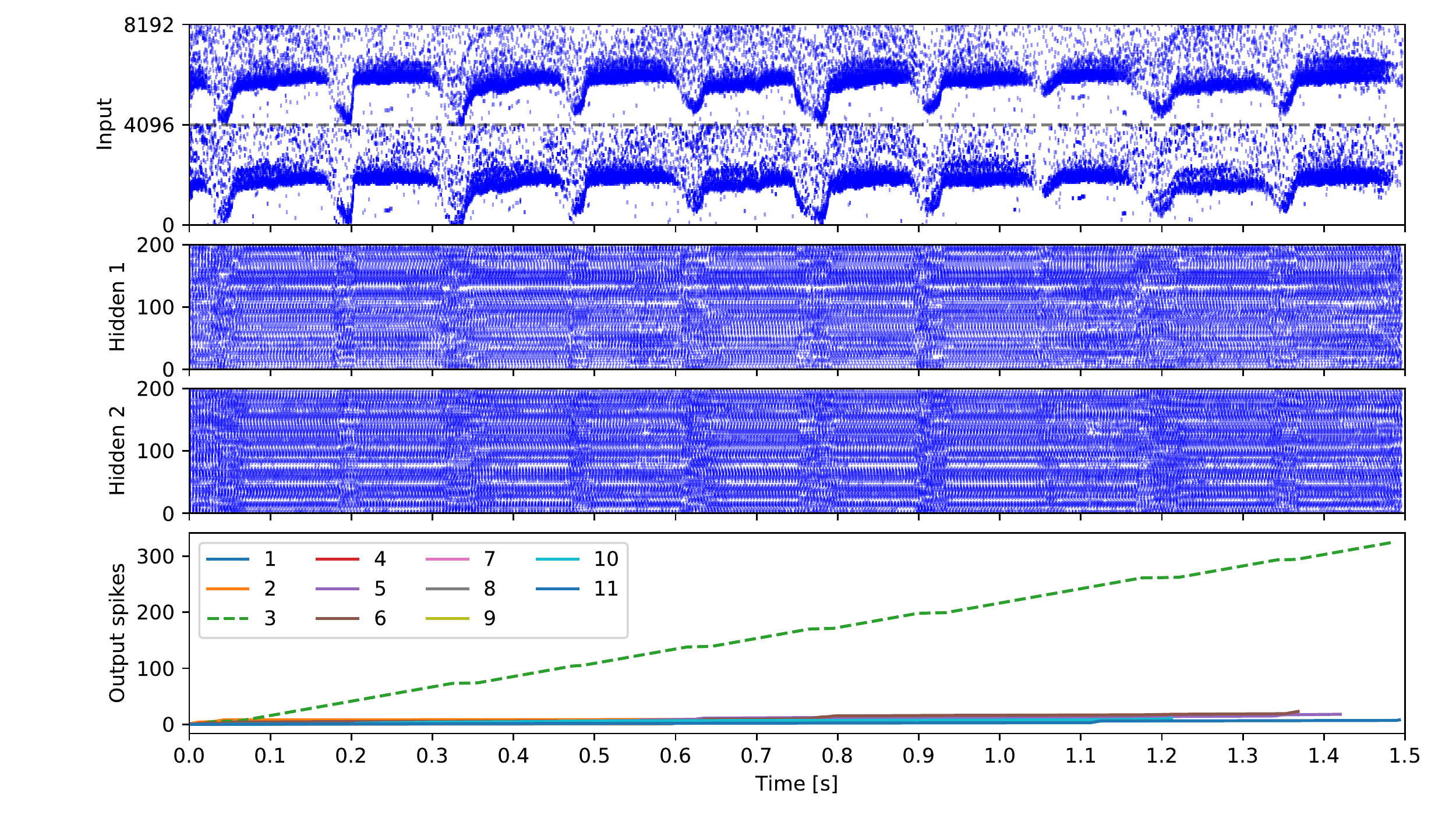}
    \end{center}
    \caption{
      Spiketrains and classification results for a test sample of class ``left hand waving'' from DvsGesture dataset.
      The network manages to correctly classify the sample in less than \SI{0.1}{\second}, with increasing confidence over time.
      The rhythm of the ``left hand waving'' motion is clearly visible in the input spiketrain.
      The neurons in the hidden layers spike close to their maximum frequency, as limited by the refractory period.
    }
  \label{fig:gesture-spiketrain}
\end{figure*}


Our evaluation on DvsGesture shows that \ac{eRBP} efficiently learns to classify motions from raw event streams with the attention mechanism introduced in \cref{sec:attention}.
The accuracy of 92.7\% is achieved after only 60 epochs, corresponding to approximately 127h of training data, see \cref{fig:exp78_77_79:acc_hist}.
This accuracy is comparable to state-of-the-art deep networks (IBM EEDN \cite{Amir2017}, 94.49\%) and other synaptic learning rules taking temporal dynamics into account (DCLL \cite{kaiser2018synaptic}, 94.19\%), both relying on convolutions.
Without the attention mechanism, the accuracy of the network drops to 86.1\%.
These results confirm that our simple covert attention mechanism provides translation invariance without convolutions, at a low computational cost.
Additionally, unambiguous samples are classified in under \SI{0.1}{\second}, with an increasing confidence over time, see \cref{fig:gesture-spiketrain}.

Since DvsGesture is a classification task, not accounting for neural temporal dynamics in the learning rule (\cref{eq:lr}) does not impact the performance much.
Indeed, the target output signal as encoded by label neurons is constant across a training sample for durations of several seconds, see \cref{fig:sample_lengths}.
We expect this omission to decrease performance significantly for a temporal regression task -- such as learning a time sequence -- where the temporal dynamics of the target signal is relevant.

\subsection{Grasp-type Recognition}

\begin{figure}[!htbp]
\centering
\includegraphics[width=0.685\columnwidth]{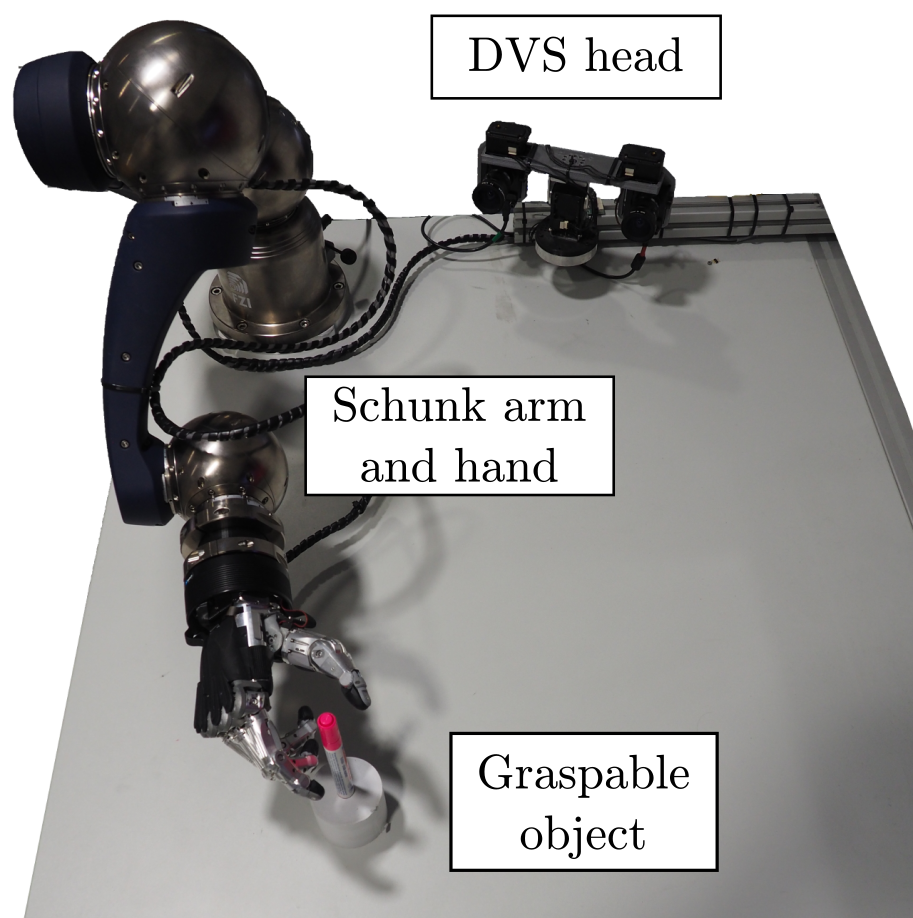}
\caption{
  Real-world grasp-type Recognition experiment setup integrating a Schunk LWA4P arm equipped with a five-finger Schunk SVH gripper and a \ac{DVS} head.
  The \ac{DVS} head performs microsaccadic eye movements to sense event streams from static scenes.
  We recorded a small four-classes dataset (ball, bottle, pen, background) of 20 samples per class.
  At test time, the detected grasp-type triggers the corresponding predefined reaching and grasping motion.
}
\label{fig:grasp}
\end{figure}

\begin{figure}[!htbp]
    \begin{flushleft}
      \begin{tabular}{@{}cccc@{}}
        ball & bottle & pen & background \\
        \includegraphics[width=0.21\columnwidth]{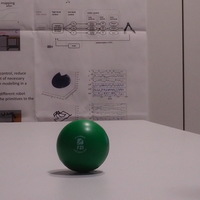} &
        \includegraphics[width=0.21\columnwidth]{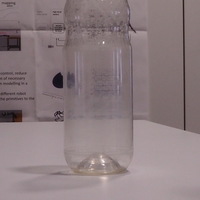} &
        \includegraphics[width=0.21\columnwidth]{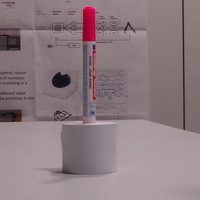} &
        \includegraphics[width=0.21\columnwidth]{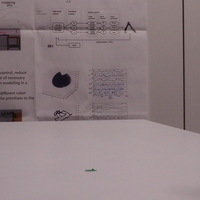} \\
        \includegraphics[width=0.21\columnwidth]{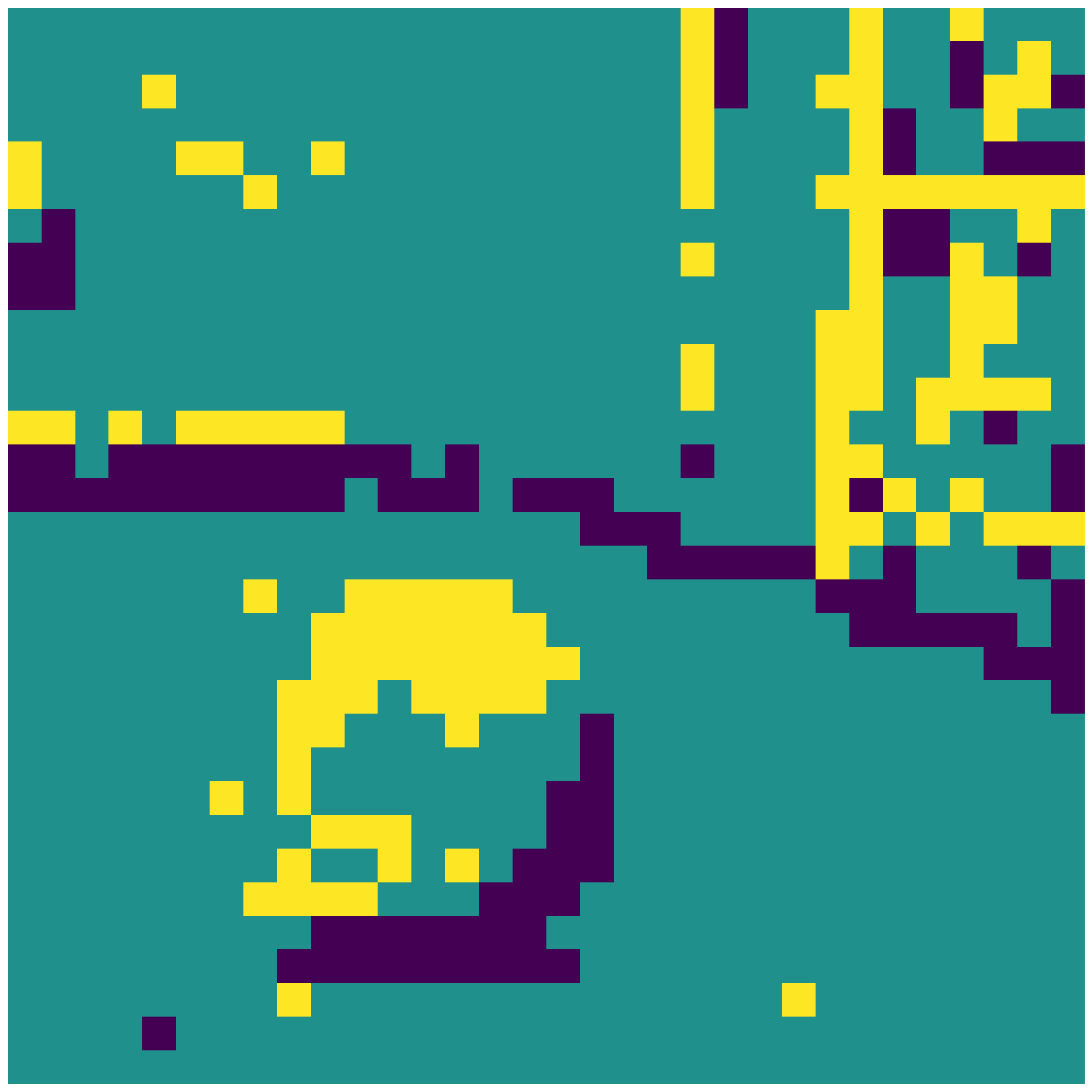} &
        \includegraphics[width=0.21\columnwidth]{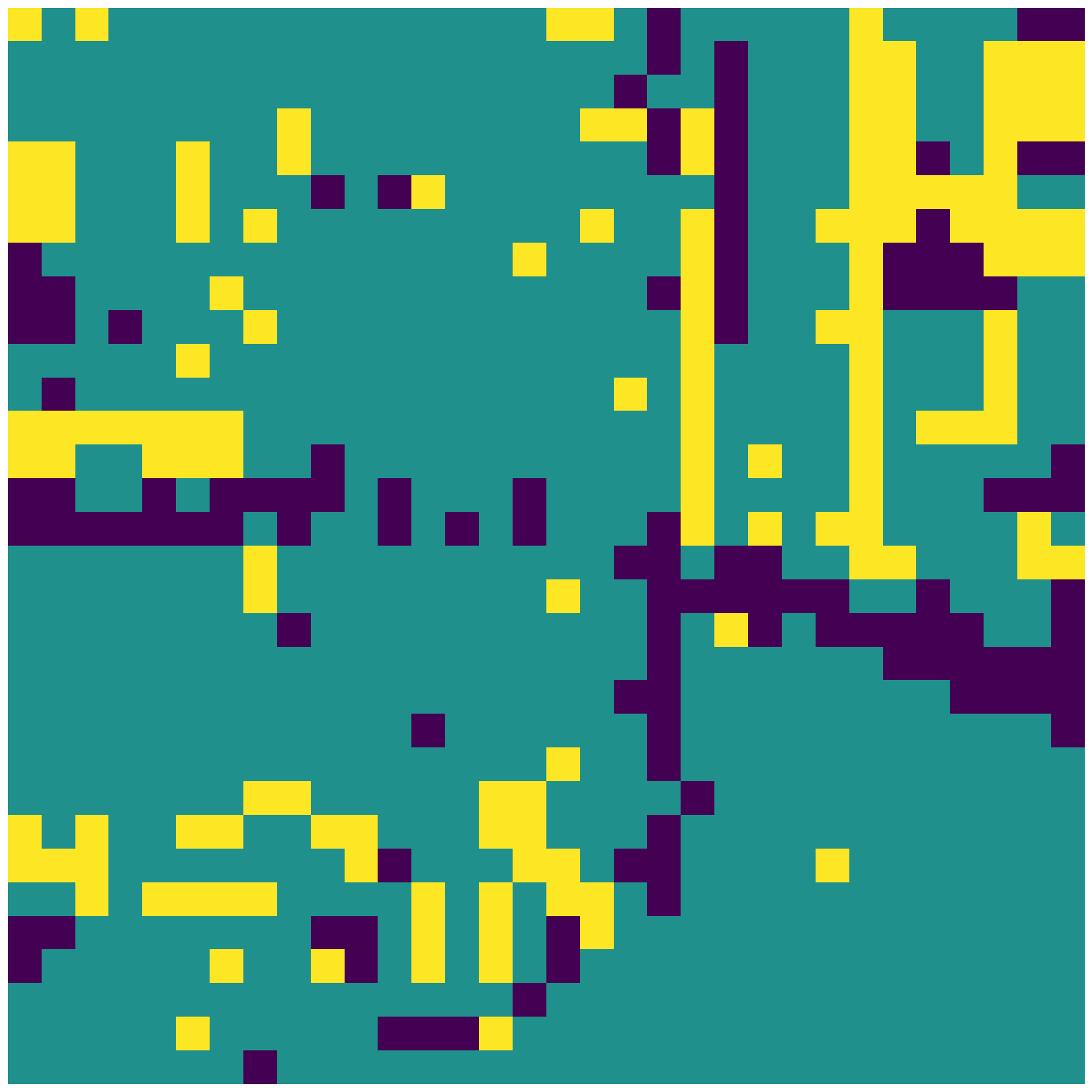} &
        \includegraphics[width=0.21\columnwidth]{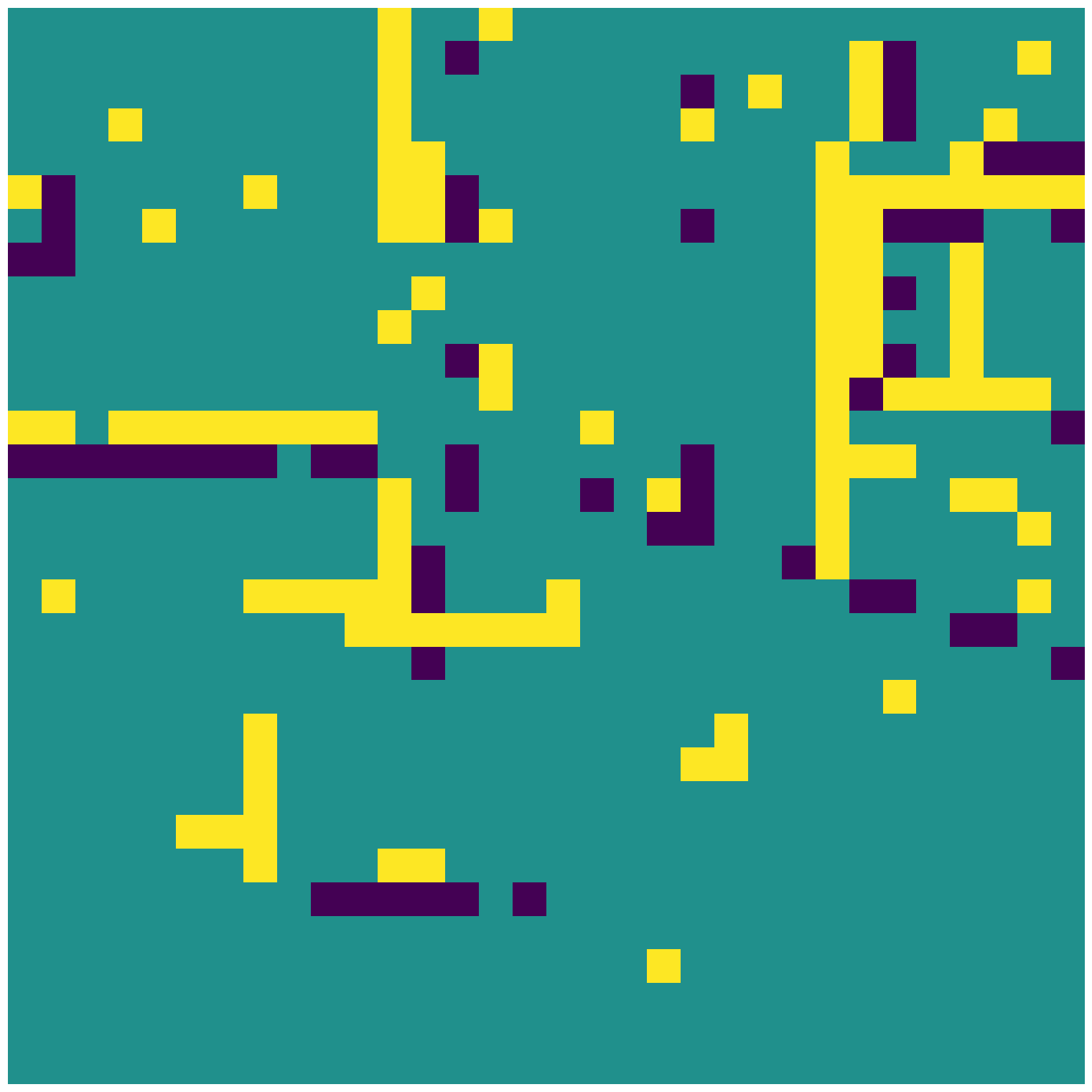} &
        \includegraphics[width=0.21\columnwidth]{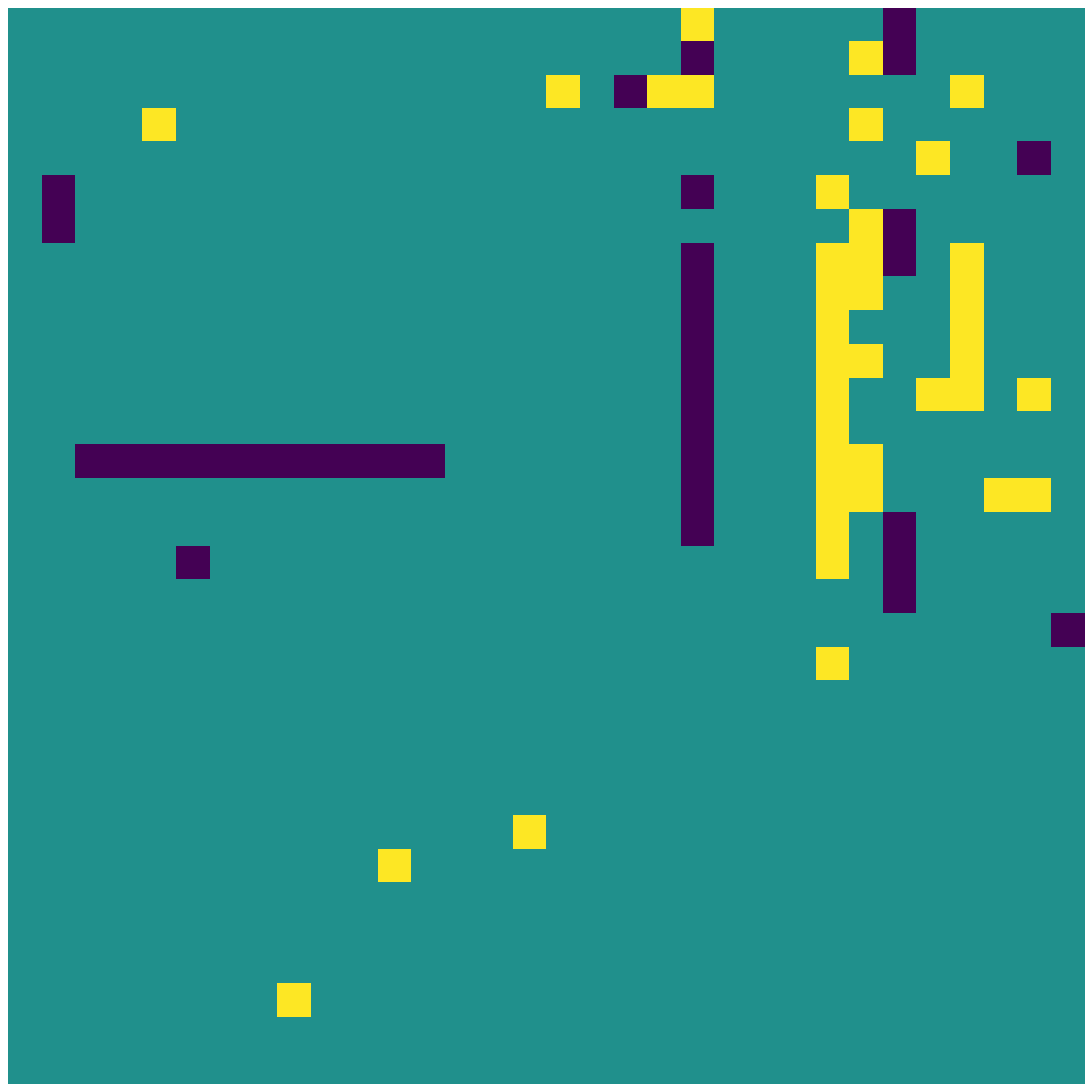} \\
        \includegraphics[width=0.21\columnwidth]{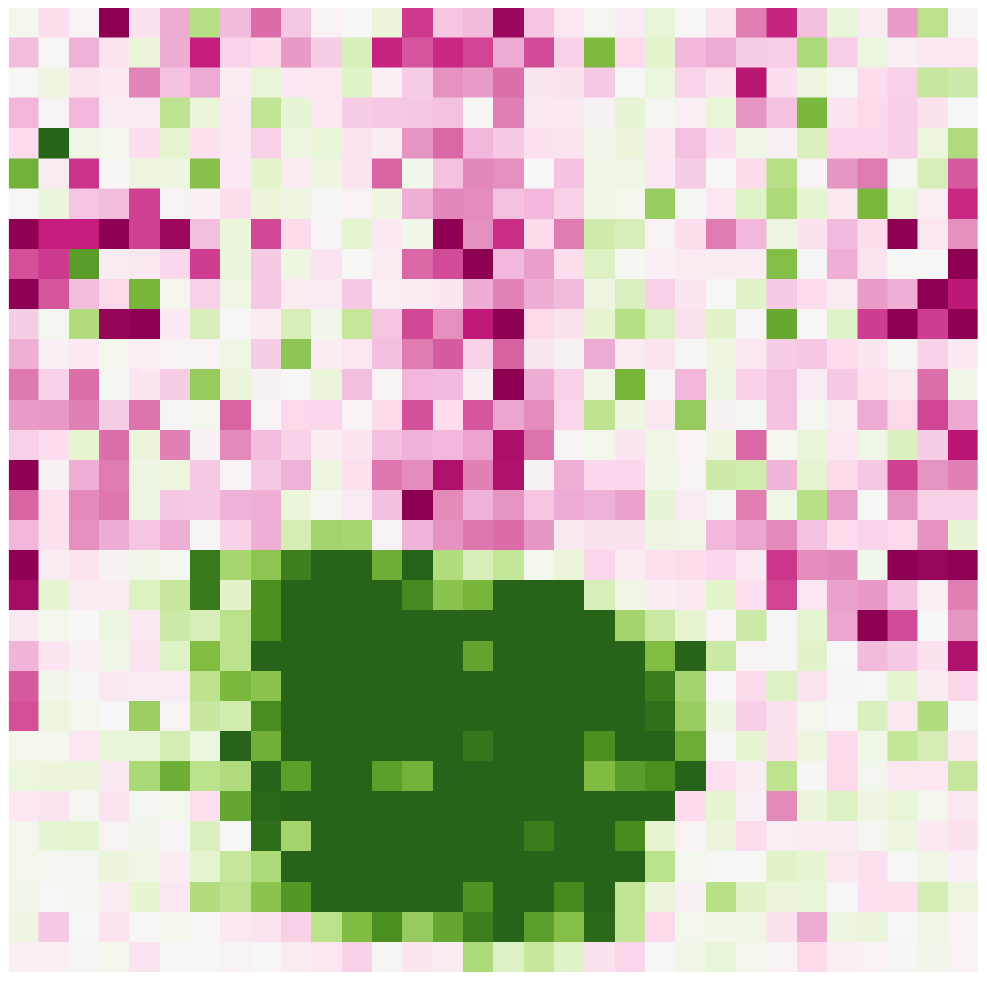} &
        \includegraphics[width=0.21\columnwidth]{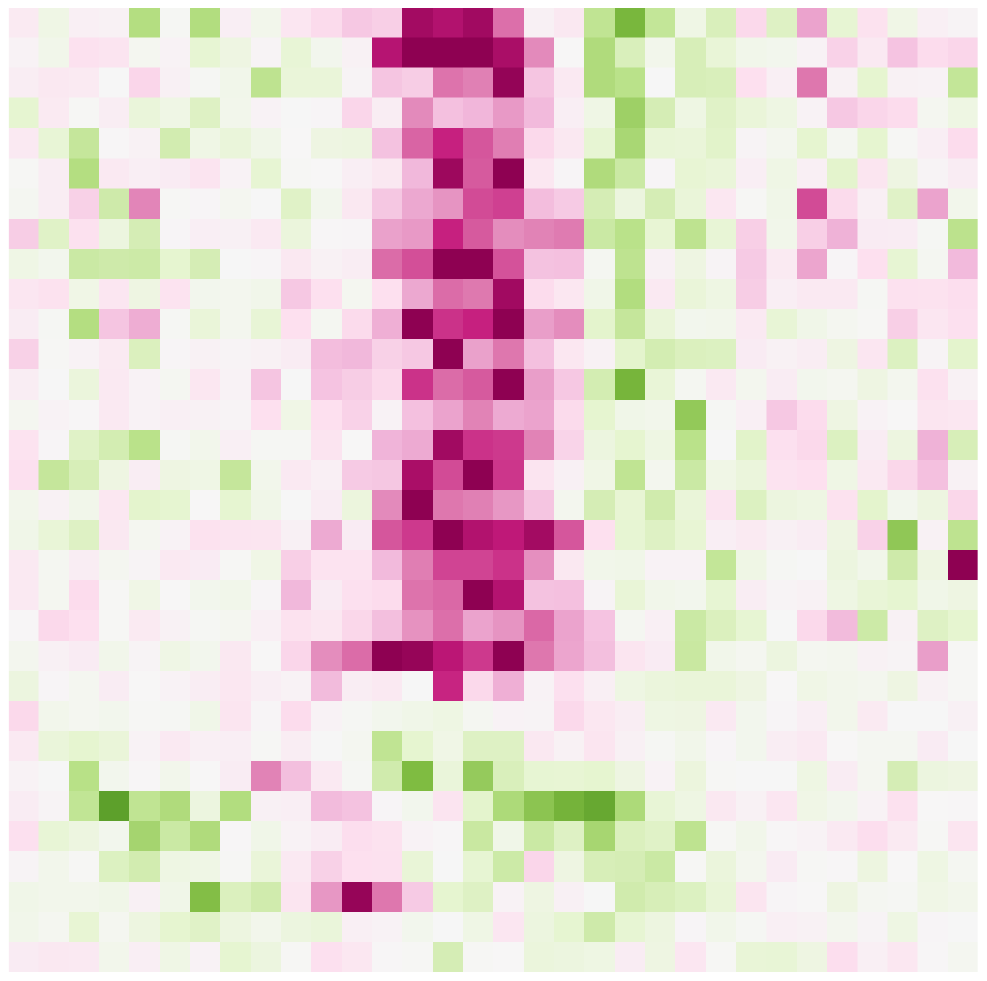} &
        \includegraphics[width=0.21\columnwidth]{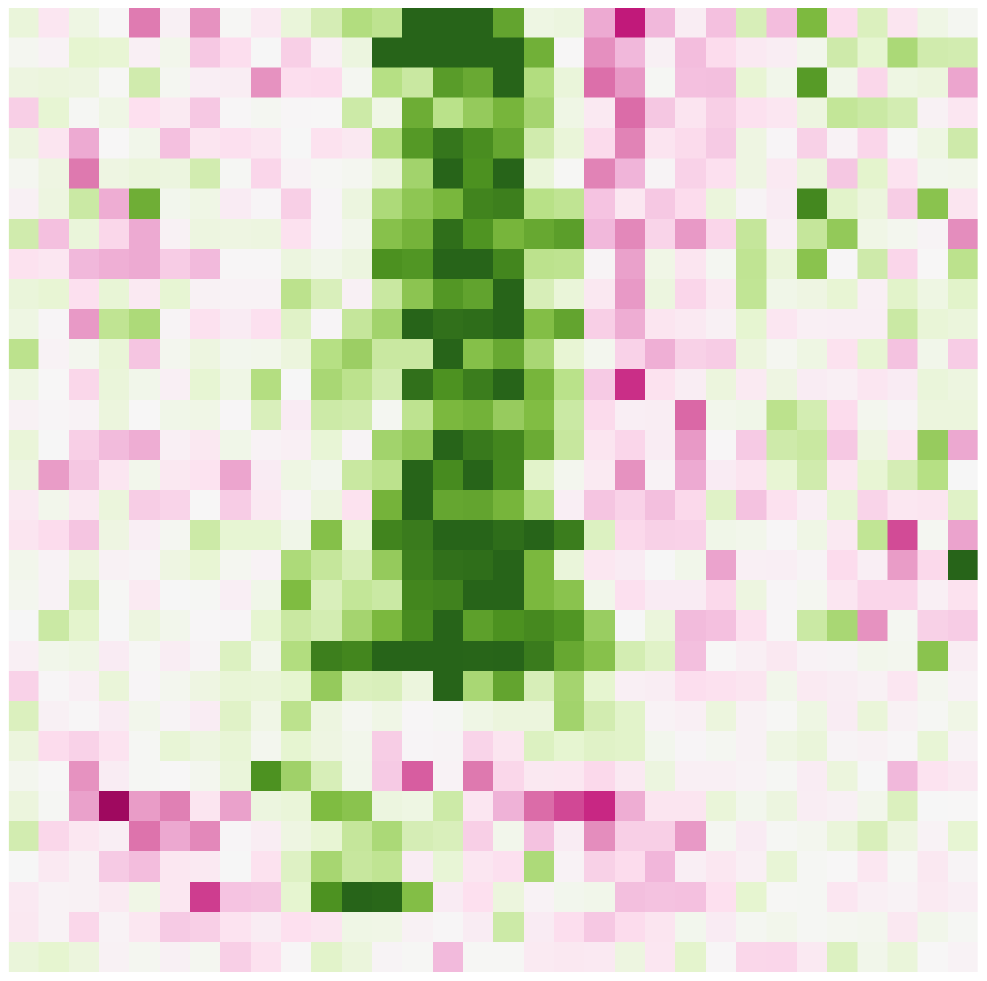} &
        \includegraphics[width=0.21\columnwidth]{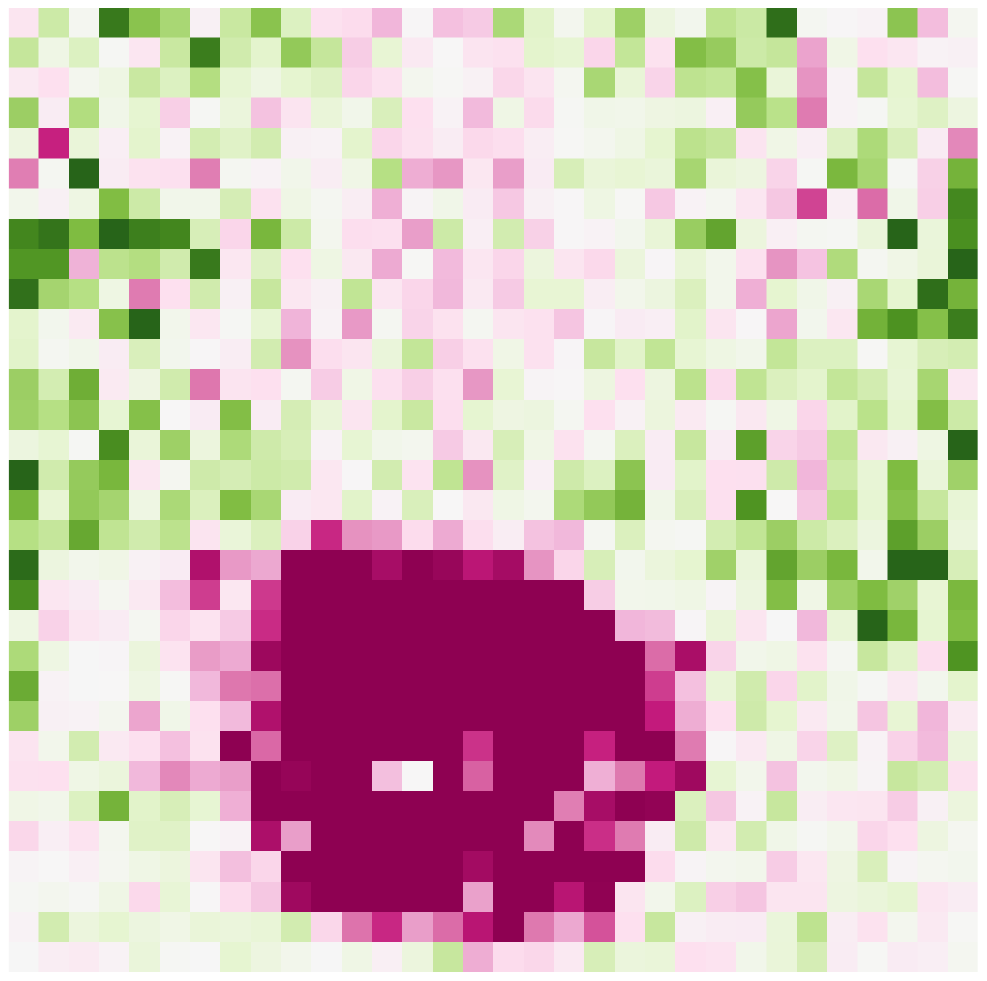}
      \end{tabular}
    \end{flushleft}
    \caption{
      Example samples and learned weights for the grasp-type recognition experiment.
      Top row: camera image of the objects.
      Middle row: integration of the address events during 15ms after microsaccade onset.
      Bottom row: projection of the synaptic weights of our 4-layers network for each label neuron onto the input after training.
      Green denotes excitation (positive influence) and pink denotes inhibition (negative influence).
    }
  \label{fig:grasp-data}
\end{figure}

In this experiment, we embody \ac{eRBP} in the real-world grasping robotic setup depicted in \cref{fig:grasp}.
In this setup, the spiking network is trained to recognize four labels corresponding to four different types of grasp: ball-grasp, bottle-grasp, pen-grasp or do nothing \cite{kaiser2017rbm}.
During training, an object of a particular class is placed on a table at a specific position.
The robotic head performs microsaccadic eye movements (similar to the N-MNIST dataset \cite{Orchard2015a}) to extract visual information from the static object.
The event streams are recorded together with the corresponding object affordance.
In this experiment, the attention window has dimension $32 \times 32$ and is fixed to match the position of the objects on the table, see \cref{fig:grasp-data} for example samples.
During testing, a microsaccade is performed and the detected object affordance triggers the adequate predefined reaching and grasping motion on a Schunk LWA4P arm equipped with a five-finger Schunk SVH gripper.
This demonstrator was implemented with the ROS Framework \cite{quigley2009ros} and the ROS \ac{DVS} driver introduced in \cite{mueggler2014event}.

\begin{figure*}[!htbp]
    \begin{center}
      \begin{tabular}{@{}cc@{}}
        Network activity for ``ball'' & Network activity for ``bottle'' \\
        \includegraphics[width=1\columnwidth]{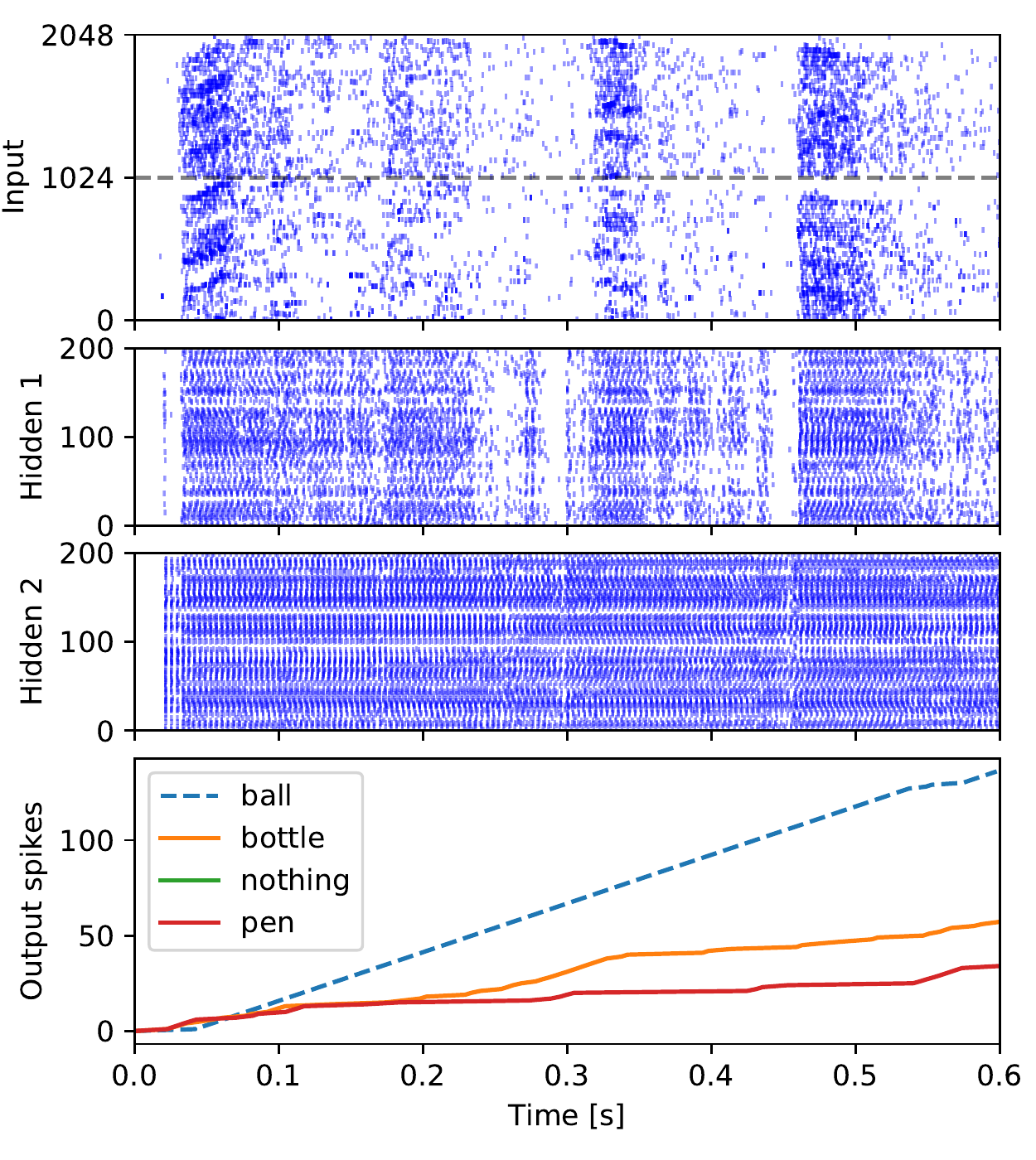} &
        \includegraphics[width=1\columnwidth]{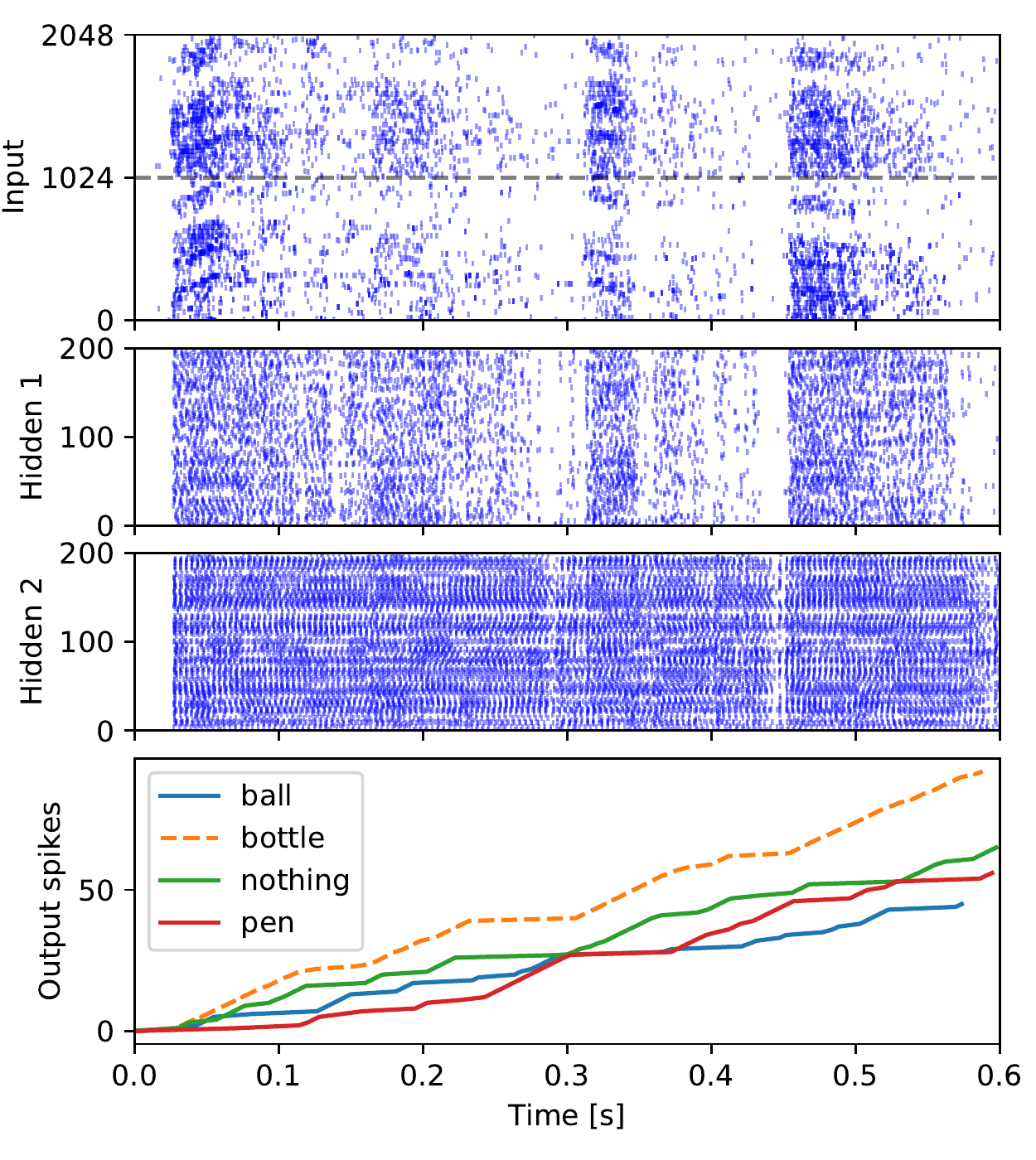} \\
        Network activity for ``pen'' & Network activity for ``background'' \\
        \includegraphics[width=1\columnwidth]{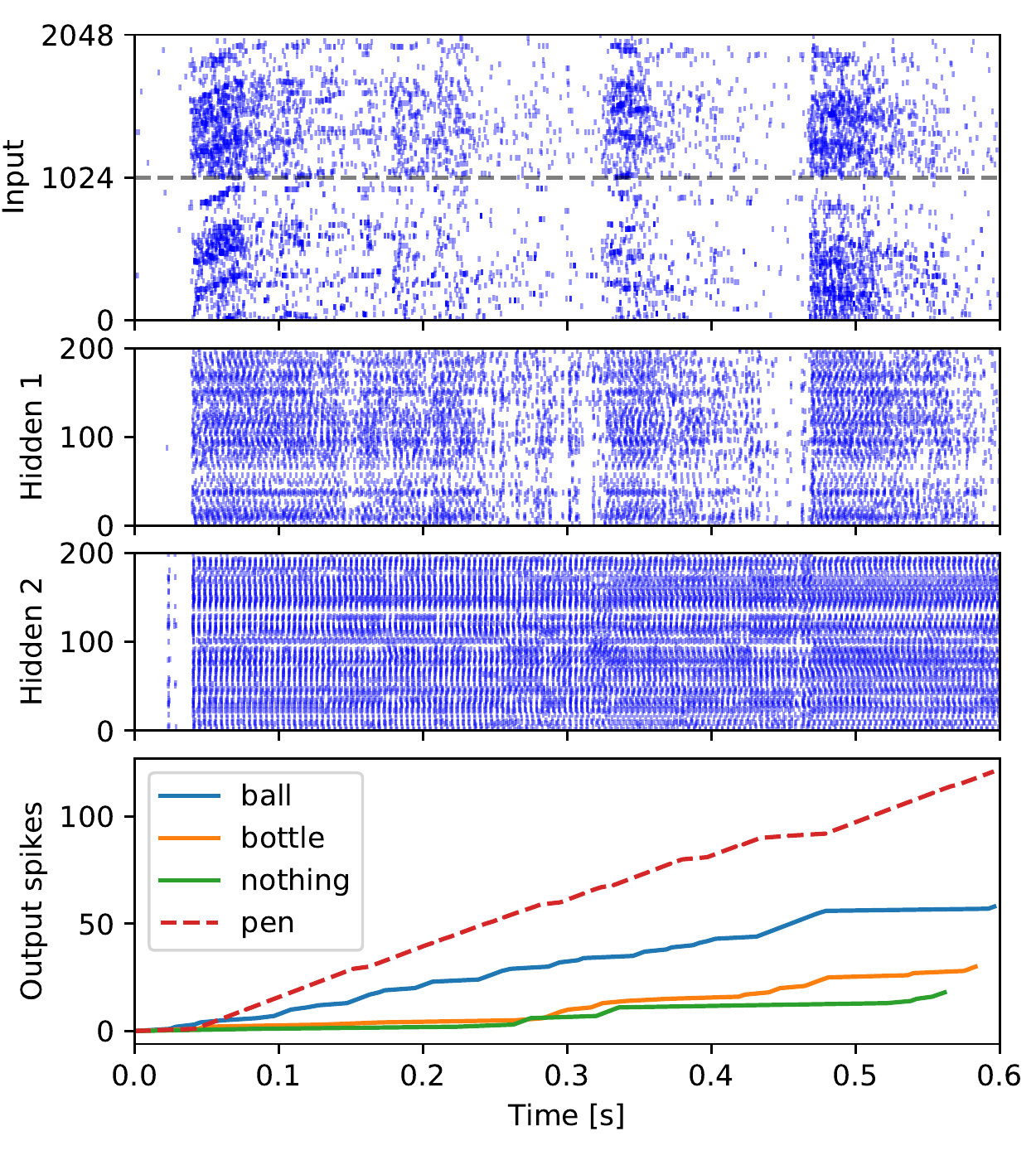} &
        \includegraphics[width=1\columnwidth]{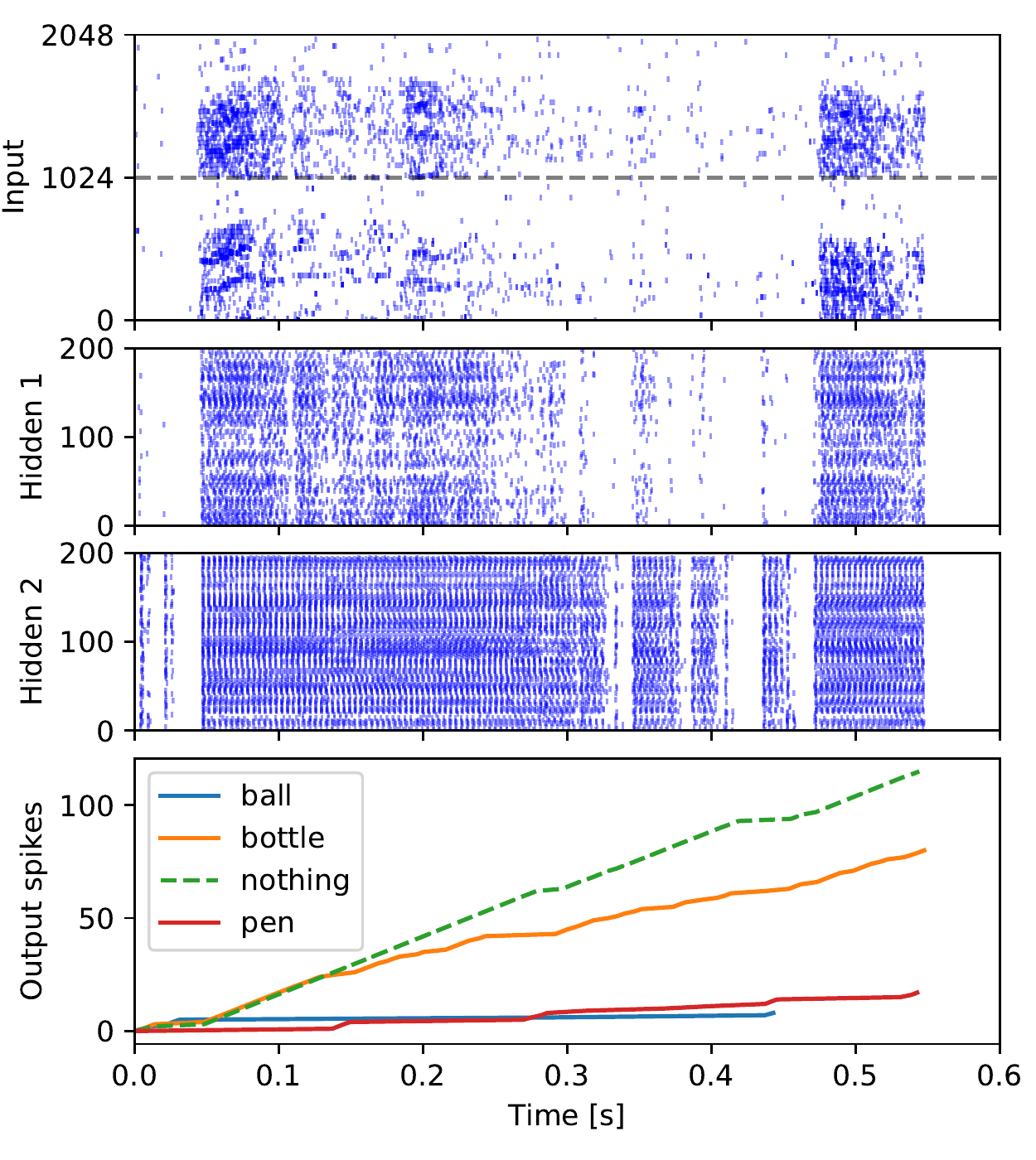} \\
      \end{tabular}
    \end{center}
    \caption{
      Spiketrains and classification results for four test samples of our grasp-type dataset.
      The network manages to correctly classify the four test samples.
      The ball and the pen are easily classified despite the small amount of training data.
      However, the transparent bottle generates few events, yielding to more uncertainty in the classification with the background.
      The three phases of the microsaccadic motion are clearly visible in the input spiketrains (first row of each plot), see \cref{sec:microsaccade}.
      However, the activity of the hidden layers does not drop instantaneously even when the input is sparse, indicating a form of short-term memory induced by the neural dynamics.
      The neurons in the hidden layers spike close to their maximum frequency, as limited by the refractory period.
    }
  \label{fig:grasp-test}
\end{figure*}

With only 20 samples per class and 30 training epochs, the network was capable of learning the four visual affordances, see the supplementary video\footnote{https://neurorobotics-files.net/index.php/s/sBQzWFrBPoH9Dx7}.
Example spiketrains and classification results at test time are shown in \cref{fig:grasp-test}.
The network classifies the ball and the pen with high confidence, but with low confidence for the bottle and the background.
This is due to the fact that we trained with a transparent bottle, generating only a few events, thus visually similar to the background.

While the microsaccade leads to sparse activity in the input layer, both hidden layers have strong, sustained activity (see \cref{fig:grasp-test}).
This indicates some form of short-term memory emerging from neural dynamics, as the network is capable of remembering the object despite receiving no events between two microsaccadic phases.
The neurons in the hidden layers spike close to their maximum frequency, as limited by the refractory period.
This is not surprising, since we did not use a regularization term to enforce sparsity in the learning rule and labels were encoded into spikes with rate coding, see \cref{sec:network}.
Except for the background class, counting the spikes at the output leads to successful classification about 100ms after microsaccade onset, comparable to human performance reported in behavioral study \cite{martin2018zapping}.

Since the \ac{DVS} does not sense colors, the network only relies on shape information, crucial for affordances.
This allowed the network to moderately generalize despite the small amount of training samples.
Indeed, a single object per affordance was used during training, but the network could recognize different objects of the same shape.
Recognition also worked when the objects were slightly moved from the reference point used for grasping.
However, the network was not robust to change in background or unexpected background motions happening during the microsaccade.
This is due to the background being learned as an additional class for the ``do nothing'' affordance.

\section{Conclusion}\label{sec:conc}

Neuromorphic engineering technology enables the design of autonomous learning robots operating at high speed for a fraction of the energy consumption of current solutions.
Until recently, the advantages of this technology were limited due to the lack of efficient learning rules for spiking networks.
This bottleneck has been addressed since the realization that gradient backpropagation can be approximated with spikes.
In this paper, we demonstrated the ability of \ac{eRBP} to learn from real-world event streams provided by a \ac{DVS}.
First, with the addition of a simple covert attention mechanism, we have shown that \ac{eRBP} achieved comparable accuracy as state-of-the-art methods on the DvsGesture benchmark \cite{Amir2017}.
This attention mechanism improved performance by a large margin in comparison to classical rescaling approaches, by providing translation invariance at a low computational cost compared to convolutions.
Second, we integrated \ac{eRBP} in a real-world robotic grasping experiment, where affordances are detected from microsaccadic eye movements and conveyed to a robotic arm and hand setup for execution.
Our results show that correct affordances are detected within about 100ms after microsaccade onset, matching biological reaction time \cite{martin2018zapping}.

For future work, \ac{eRBP} could be replaced by more recent learning rules accounting for neural temporal dynamics \cite{neftci2017event,Zenke_Ganguli17_supesupe,kaiser2018synaptic,bellec2019biologically}.
This would enable the setup to be extended to temporal sequence learning and reinforcement learning tasks \cite{tieck2018learning,kaiser2019spore}.
Additionally, other components of the grasp-type recognition experiment could be implemented with spiking networks, such as reaching motions \cite{tieck2019combining,tieck2018controlling}, grasping motions \cite{tieck2017towards} and depth perception \cite{kaiser2018microsaccades}.
This work paves the way towards the integration of brain-inspired computational paradigms into the field of robotics.

\section*{Acknowledgment}
This research has received funding from the European Union's Horizon 2020 Framework Programme for Research and Innovation under the Specific Grant Agreement No. 785907 (Human Brain Project SGA2).

\balance






\end{document}